\documentclass[review]{elsarticle}
\usepackage{hyperref}
\usepackage{float}
\usepackage{verbatim} %comments
\usepackage{apalike}
\restylefloat{figure}
\restylefloat{table}

\usepackage{amsmath}
\usepackage{algorithm}
\usepackage{algorithmic}
\usepackage{graphicx}
\usepackage{mathtools}
\journal{Expert Systems with Applications}

\biboptions{authoryear}

\begin{document}

\begin{frontmatter}

\begin{titlepage}
\begin{center}
\vspace*{1cm}

\textbf{ \large Video Violence Recognition and Localization Using a Semi-Supervised Hard Attention Model}

\vspace{1.5cm}

% Author names and affiliations
Hamid Mohammadi$^a$ (hamid.mohammadi@aut.ac.ir), Ehsan Nazerfard$^a$ (nazerfard@aut.ac.ir) \\

\hspace{10pt}

\begin{flushleft}
\small  
$^a$ Department of Computer Engineering and Information Technology, Amirkabir University of Technology, Valiasr Sq., 350 Hafez Ave. Tehran, Iran

\vspace{1cm}
\textbf{Corresponding Author:} \\
Ehsan Nazerfard \\
Department of Computer Engineering and Information Technology, Amirkabir University of Technology, Valiasr Sq., 350 Hafez Ave. Tehran, Iran \\
Tel: (+98) 6454-2707 \\
Email: nazerfard@aut.ac.ir

\end{flushleft}        
\end{center}
\end{titlepage}

\title{Video Violence Recognition and Localization Using a Semi-Supervised Hard Attention Model}

\author[label1]{Hamid Mohammadi}
\ead{hamid.mohammadi@aut.ac.ir}

\author[label1]{Ehsan Nazerfard \corref{cor1}}
\ead{nazerfard@aut.ac.ir}

\cortext[cor1]{Corresponding author.}
\address[label1]{Department of Computer Engineering and Information Technology, Amirkabir University of Technology, Valiasr Sq., 350 Hafez Ave. Tehran, Iran}

\begin{abstract}
%\hl
{The significant growth of surveillance camera networks necessitates scalable AI solutions to efficiently analyze the large amount of video data produced by these networks. As a typical analysis performed on surveillance footage, video violence detection has recently received considerable attention. The majority of research has focused on improving existing methods using supervised methods, with little, if any, attention to the semi-supervised learning approaches. In this study, a reinforcement learning model is introduced that can outperform existing models through a semi-supervised approach. The main novelty of the proposed method lies in the introduction of a semi-supervised hard attention mechanism. Using hard attention, the essential regions of videos are identified and separated from the non-informative parts of the data. A model's accuracy is improved by removing redundant data and focusing on useful visual information in a higher resolution. Implementing hard attention mechanisms using semi-supervised reinforcement learning algorithms eliminates the need for attention annotations in video violence datasets, thus making them readily applicable. The proposed model utilizes a pre-trained I3D backbone to accelerate and stabilize the training process. The proposed model achieved state-of-the-art accuracy of 90.4\% and 98.7\% on RWF and Hockey datasets, respectively.
}
\end{abstract}

\begin{keyword}
deep reinforcement learning \sep violence detection \sep hard attention \sep video classification \sep semi-supervised learning
\end{keyword}

\end{frontmatter}

\section{Introduction}
\label{Introduction}

% 1. Human-based vs AI-based surveillance systems
A comprehensive and dedicated violence monitoring system is becoming increasingly necessary as social unrest, social violence, and homicide cases increase. A security system's most significant weakness has always been the reliability of its human agents \citep{gill2005assessing, bugeja2018investigation}. The number of surveillance cameras installed worldwide is rapidly exceeding the data-load capacity of legacy human-operated surveillance and monitoring systems\citep{gill2005assessing, bugeja2018investigation}. Violence monitoring systems based on artificial intelligence have significant potential as a reliable and scalable alternative to human-based surveillance systems. As opposed to human-based surveillance systems, artificial intelligence and machine learning surveillance systems offer predictable downtime, reliability, and effortless scaling\citep{nguyen2020artificial, shidik2019systematic, sung2021design}.

% 2. Dataset gathering difficulty + RWF dataset
The quality of deep learning models depends on the quality of their data. It is difficult to obtain surveillance footage due to its security and privacy concerns. Almost all privately collected datasets in this field are covered by non-disclosure agreements, which prevent the public from accessing them, including artificial intelligence researchers. In light of this issue, it is valuable for the community to collect and publish a surveillance-based human violence dataset. The significant advantage of the RWF dataset \citep{cheng2021rwf} over other video violence datasets (Hockey fights \citep{nievas2011violence} and Movie violence \citep{gong2008detecting}) is collection of a relatively large set of real-world surveillance violence videos.

% 3. Temporal information in video classification + Temporal feature extractors
In video classification, temporal knowledge is essential \citep{yao2016spatio, murthy2014influence}. A single video frame provides little insight into human violence \citep{feichtenhofer2016convolutional}. Having temporospatial understanding requires recurrent or 3D architectures in deep neural networks \citep{feichtenhofer2016convolutional, yao2016spatio, algamdi2019learning, du2017recurrent}. In action and violence recognition research, recurrent networks, 3D convolutional layers, and multi-stream architectures are common themes\citep{zong2021motion, wang2021multi, shi2020skeleton}. However, recurrent neural networks, such as LSTMs or GRUs, cannot grasp the complete dependency between consecutive frames \citep{zeyer2019comparison, ezen2020comparison, wang2019language}. While transformers have state-of-the-art accuracy in video classification and action recognition, they lack the computational agility and performance required for cost-effective large-scale video surveillance platforms \citep{arnab2021vivit, girdhar2019video}. The 3D convolutional layers can be used to capture the temporal information in video using 3D convolutional filters with parameters sharing capabilities \citep{feichtenhofer2016convolutional, yao2016spatio, algamdi2019learning, du2017recurrent}. As a result of their effectiveness and efficiency, these layers are ideal for this study.

% 4. Auxiliary features
Video violence recognition accuracy is enhanced by the use of auxiliary features \citep{tu2018multi}. Aside from raw RGB video frames, additional features such as RGB-differences \citep{wang2017structured}, optical flow \citep{sevilla2018integration}, pose estimation \citep{luvizon20182d, pham2020unified}, and deep-learning-based features \citep{li2016vlad3, khan2018scale, xiao2019self} are utilized to capture different aspects of the input videos. When the network itself cannot comprehend temporally spatial patterns, motion estimation features, such as RGB-difference and optical flow characteristics, assist in capturing the temporal interdependence of video frames \citep{wang2018appearance}. Additionally, pose estimation, and deep-learning-based features are intended to convey contextual information about each frame. Pose estimation features represent the position and movement of the human body \citep{luvizon20182d, pham2020unified}. Deep features (extracted from a computer vision backbone network trained on tasks, such as image classification or object detection) represent the environment, background, and objects in a latent vector space suitable for action understanding and classification \citep{li2016vlad3, khan2018scale, xiao2019self}.

% 5. The cost of auxiliary features
As valuable as these features are, their extraction incurs an additional computational cost detrimental to real-world applications. The extraction of accurate optical flow requires specialized hardware or significant computational resources (equal to or exceeding the costs of the primary neural network in some cases) \citep{sun2018optical}. As a result, it is advantageous to identify more cost-effective methods of enhancing video violence identification models without adding additional features.

% 6. Main idea of paper: removing redundant information
The proposed method for improving video violence recognition relies on the premise that hard attention enhances model accuracy. Surveillance videos contain considerable information redundancy as humans are the only subjects in a pipeline for violence recognition. Therefore, the background information can be eliminated without compromising vital details\citep{sharma2015action}. By using hard attention, the video violence recognition model eliminates redundant information from the input frames. The reduction of redundancy and increased focus on the useful information in a video increases the model's accuracy by reducing its search space and avoiding overfitting.

% 7. Disadvantage of supervised hard attention *
% Supervised methods of implementing hard attention have a disadvantage compared to semi-supervised and unsupervised methods. The cost of additional annotation on a dataset makes supervised approaches inefficient. Most action recognition/violence detection datasets suffice for only the classification annotations of videos. Adding localization annotations at the frame level increases the annotation cost considerably.

% 8. Why reinforcement learning to create hard attention *
Reinforcement learning is an effective method to implement hard attention \citep{rao2017attention, driessens2019focused, shen2018reinforced, mott2019towards}. In reinforcement learning methods, partial signals (rewards) are used to optimize a global criterion \citep{sutton2018reinforcement}. A reinforcement learning implementation of the hard attention method can utilize the video level annotation information to learn the functionality of a region proposal model. Reinforcement learning methods can select a region of interest by expressing the region proposal information in action space. Furthermore, other partial information acquired from the video could provide additional learning signals for the reinforcement learning method \citep{sutton2018reinforcement}. In order to improve the precision of the reinforcement learning implementation of hard attention, information such as the region of motion, objects present in the environment, skeleton models, and similar data can be included in the reward shaping process.

% 9. Novelty of this paper
%\hl
{The novelty of this study lies in the use of semi-supervised reinforcement learning techniques to create a hard attention mechanism to improve the purity of visual data in order to achieve state-of-the-art accuracy. Unlike the mainstream state-of-the-art approaches, this method takes a reductive approach to model improvement by focusing on removing less helpful information. It is possible to apply the semi-supervised hard attention approach to the existing video violence datasets without making any modifications. The region of interest is learned based on the annotations at the video level.}

% 10. Paper organization
Accordingly, the remainder of the paper is organized as follows: Section \ref{Related Work} reviews the literature on video violence detection and reinforcement learning. Section \ref{method} introduces the proposed model, followed by the evaluation results in Section \ref{Experiments}. In Section \ref{Discussion}, the pros and cons of the proposed method are thoroughly discussed. Lastly, Section \ref{Conclusion and future work} discusses conclusions and future research.

\section{Related Work}
\label{Related Work}
% \linenumbers

% 1. The general categorization of the related work
%\hl
{The background literature can be divided into two different perspectives. In the first perspective, research is being conducted to find supervised models to classify videos based on the presence of specific actions. The second perspective involves reinforcement learning approaches to improve computer vision tasks (such as classification, detection, or tracking). The research presented here combines the perspectives mentioned above.}

\subsection{Video action classification}

% 2. Processing the third dimension + a common video classification architecture
In addition to the established image classification methods, video as three-dimensional data poses additional challenges. The addition of the third dimension requires the use of specialized features and representation learning techniques\citep{hara2017learning, wang2018human, nazir2018evaluating}. The temporal dimension can be captured using techniques such as recurrent layers  \citep{liu2016spatio, ullah2017action, majd2020correlational, liu2017global}, 3D convolutional layers \citep{ji20123d, yang2019asymmetric, zhou2018mict, hara2017learning}, and, more recently, transformers \citep{plizzari2021spatial, li2021trear, mazzia2021action, girdhar2019video}. Recurrent layers have a reputation for inconsistent training and poor temporal learning \citep{vaswani2017attention}; however, 3D convolutional layers and transformers are highly effective in this field. Due to this, most cutting-edge techniques use 3D convolutional networks and transformers to map temporal information to latent features. Such networks provide the backbone for the extraction of features in a video classification model. The backbone is followed by a simple classifier, i.e., a fully connected neural network, to form an end-to-end video classification model.

% 3. Simple and complicated temporal features
A temporal feature may be as simple as the difference between consecutive frames or more complex such as optical flow. Using richer forms of temporal data (e.g., optical flow) will result in more accurate models \citep{sevilla2018integration, sun2018optical}. Conversely, the trade-off between accuracy and performance leads to the use of superficial temporal features (such as RBG-difference) in some applications \citep{zhang2016rgb, hu2018deep, crasto2019mars, wang2018cooperative}.

% 4. Auxiliary features and their computational cost
The accuracy of action recognition can be improved by including additional explicit information in addition to RBG frames and motion features. Visual cues \citep{tu2018semantic, wang2016mofap} and skeleton estimations \citep{yan2018spatial, plizzari2021spatial, shi2019skeleton} are examples of such data. In spite of the increased accuracy, each additional data type adds two overheads to the performance of the activity recognition system. Firstly, each feature requires additional computation during extraction. As a result of the substantial processing power required for extracting motion vectors from RGB frames, researchers have removed, replaced, or estimated this feature \citep{sun2018optical, wang2013action}. Secondly, each additional data stream makes the neural network larger and more expensive to compute \citep{simonyan2014two}.

% 5. Two-stream architecture
The integration of additional data types beyond the RGB video frames has created a unique architecture for action recognition models. Two-stream and multi-stream neural networks utilize neural networks with two or more backbones that can accept multiple types of data. Following the extraction of features from each backbone, the features are fused and classified. This architecture is common in video action recognition approaches \citep{simonyan2014two, tu2018multi, shi2020skeleton, wang2021multi}.

\subsection{RL-based attention}

% 1. Data redundancy in videos + why attention helps
Videos contain a great deal of redundant information. In most cases, categorizing a video based on human action does not require understanding what is taking place in the background. A neural network's accuracy can be improved by removing excessive information and purifying the data \citep{song2022learning}. Accordingly, the accuracy of action recognition can be improved through the use of soft or hard attention in network architectures.

% 2. Soft attention definition
Soft attention can be thought of as a weighted version of the original data. Weights for each data region are automatically learned during the neural network training process \citep{sharma2015action, liu2017global, song2017end, li2020spatio}. A general form of soft attention is represented in \autoref{eq:softattention}.

\begin{equation}
\label{eq:softattention}
\overset{\mathclap{\text{attention output}}}{\overset{\Biggl|}{c_{i}}} = \sum_{j=1}^{\overset{\mathclap{\text{Total number of inputs}}}{\overset{\Biggl|}{T}}} \underbrace{softmax(\overset{\mathclap{\substack{\text{hidden state} \\ \text{step i-1}}}}{\overset{\Biggl|}{s_{i-1}}} . h_{j})}_{\text{attention coefficient}} . \underbrace{h_{j}}_{\text{input index j}}
\end{equation}

% 3. Soft attention formula description
%\hl
{A sequence of input vectors is represented by $h$. At each step of the soft attention function ($c_{i}$), the output is the weighted sum of the input matrices. The weight of each input matrix ($h_{j}$) represents the attention coefficient calculated at each step for each input matrix. At the current index, the coefficient depends on the input matrix at the current index and the hidden state of the attention layer at the previous index ($s_{i}$).}

% 4. Hard attention definition, supervised and semi-supervised
% Why no hard attention formula here?
Hard attention is a binary form of soft attention in which zero-weighted values are completely omitted from the input. As a result, the output size of a hard attention function is not continuous and may differ from its input size. It is possible to implement hard attention using supervised, self-supervised, and semi-supervised techniques. The implementation of hard attention using supervised methods is costly since the localization annotations must be manually entered into the dataset. It is also possible to implement hard attention through self-supervised methods \citep{manchin2019reinforcement}. For example, as most actions are considered a type of motion, removing motionless parts of a video is a form of hard attention \citep{crasto2019mars}. Motion attention guarantees the inclusion of most activities in the attention area (except for activities with no motion, such as sleeping, sitting, pointing, and so on); however, it creates a lot of redundant data. Using motion attention methods, it is difficult to distinguish a particular action occurring in a busy street from other movements in the video.

% 5. Spatial transformers + their drawback
Spatial transformer networks (STNs) are end-to-end solutions to implement hard attention in neural networks \citep{jaderberg2015spatial}. STNs are neural layers that compute matrix spatial transformations (for example, scaling, rotating, and cropping). These networks implicitly learn the transformation required for each input according to the global accuracy of the larger neural network. With a few minute design changes, convolutional neural networks (CNNs) can be transformed into hard attention CNNs utilizing STNs \citep{li2018harmonious, malinowski2018learning}. However, since their output is simply a modified version of the input image, these networks are constrained by the resolution of the input image. In order to gain the maximum benefit from the high-resolution input, it is necessary to utilize alternative methods.

% 6. How to implement RL hard attention
Classification, localization, and tracking performance are good reward sources for reinforcement learning algorithms. As a result, the main objective is to assign the reinforcement learning model the task of improving the accuracy of the underlying model. Reinforcement learning models improve accuracy by focusing on important information in videos, given their attention ability. This approach eliminates the need for extra information (e.g., annotations) in the learning data. The use of neural networks in reinforcement learning models (or deep reinforcement learning) allows advanced computer vision algorithms to be applied to reinforcement learning. As a result, state-of-the-art computer vision neural architectures are paired with reinforcement learning techniques to improve performance on current tasks. A combination of these approaches is investigated in computer vision tasks, including object localization \citep{wang2018multitask, jie2016tree, caicedo2015active, mathe2016reinforcement}, and visual tracking \citep{luo2019end, ren2018collaborative, yun2018action, zhong2019decision, cui2021remote}. For a more innovative example, \citet{rao2017attention} used reinforcement learning in order to create temporal attention in the face recognition task. Identifying faces that are most likely to be correctly recognized in a timeline is an ideal task for deep reinforcement learning.

% 7. Examples of RL in CV
% Hard attention can be a spatial, temporal, or Spatio-temporal sampling of information. Region proposal in an object detection task is an example of spatial hard attention that is done using reinforcement learning \citep{pirinen2018deep, uzkent2020efficient}. \citet{rao2017attention} have used reinforcement learning to create temporal attention in the face recognition task. Selecting faces in a timeline that are most likely to be correctly recognized is a fit task for deep reinforcement learning hard attention.

% Different implementations of hard attention mechanism
% Finding the perfect information in Spatio-temporal data combines the previous ideas to propose an attention mechanism suitable for video \citep{wang2020attentionnas, dong2019attention}. However, there are several ways to design attention mechanisms, particularly hard attention. The attention mechanism may freely choose any spatiotemporal region or be restricted to certain areas. A multi-step or one-step process might be used to choose the focus area. The answer to such questions depends on the type of task, training data, and use case.

\section{Method}
\label{method}
% \linenumbers

% 1. This section describes the model, its interactions, and its training
%\hl
{A reinforcement learning model and a set of strategies for describing the observations and actions taken by an agent to detect violence in videos are presented in this study. Through the addition of hard attention and semi-supervised learning capabilities, the deep reinforcement learning agent improves the established deep video classification models. In order to train the semi-supervised model, the dataset annotations are converted into reward signals for the agent. The deep reinforcement learning model is improved by reward shaping and train stabilization.}

\subsection{Semi-Supervised Hard-Attention}
\label{3:1}

% 2. Base assumptions of hard attention
SSHA, short for \textbf{S}emi-\textbf{S}uprevised \textbf{H}ard \textbf{A}ttention, is based on two assumptions: (i) Given a high-quality dataset, a neural network with a larger size performs better than a similar neural network with smaller size. (ii) Removing redundant information from neural network input results in either a smaller model (fewer parameters) with roughly the same accuracy or a model of the same size (same number of parameters) the same size but with higher accuracy.

% 3. Resize info loss + information redundancy
For the purpose of reducing the computational cost, input images are often shrunk to a smaller size. The information contained in an image is lost as it becomes smaller because details are removed, and key characteristics are represented in a smaller vector space (the number of pixels is reduced). Consequently, it is common in the computer vision community to propose a range of models with different input sizes (thus varying the number of parameters) in order to address the trade-off between accuracy and computational cost. The redundancy of information is also a common problem in computer vision applications. It is critical to note that redundant data increases the dimensionality of input data. This fills the input data with information that could have been used to demonstrate valuable features. In action recognition tasks, data redundancy is evident since the majority of information is redundant in each video (i.e., environment, background objects, almost everything except the subject of the action).

% 4. Combining our assumptions using hard attention
The proposed method combines the above assumptions with minimal drawbacks. Hence, obtaining high accuracy and performance with a smaller model and fewer redundant inputs. With hard attention, redundant data can be removed from a network's input, improving accuracy without the need for a larger network. A greater amount of computational power is available to process the valuable information once redundant information is removed.

% 8. Generalizability of hard attention
%\hl
{Compared to using the auxiliary features, the hard-attention methodology is more generalizable and cost-effective. In addition to the computational overhead of auxiliary features, some features, such as skeleton estimation, are selected based on the video violence detection application (importance of human body dynamics). Due to their application-specific nature, such features cannot be used in a broader range of computer vision applications. The low computational overhead and application-neutral assumptions of the hard attention mechanism make it suitable for a wider range of applications in computer vision.}

% 5. Hard attention form in computer vision
The concept of hard attention in computer vision can be understood as the process of cropping out redundant information from each frame. As a result, the hard attention task can be formulated as a method of determining the coordinates of a crop function. \autoref{eq:hardattention} is an interpretation of such an approach.

\begin{equation}
\label{eq:hardattention}
\overset{\mathclap{\text{attention output}}}{\overset{\Biggl|}{c}} = \overset{\mathclap{\text{crop function}}}{\overset{\Biggl|}{f_{crop}}}(\underset{j}{arg\,max}( \overbrace{f_{score}}^{\mathclap{\substack{\text{attention score function} \\ \text{DRL network}}}} ( \underset{\mathclap{\substack{\text{input at region j}}}}{\underset{\Biggl|}{h_{j}}} )), h)
\end{equation}

% 6. Description of hard attention equation
\autoref{eq:hardattention} %\hl
{is centered on deep reinforcement learning (DRL) attention scoring model ($f_{score}$). The DRL model is a function that scores regions based on their importance in the detection of violence ($h_{j}$). The region with the maximum attention score is then cropped using the crop function ($f_{crop}$) and considered to be the output of hard attention. The crop function is a light image processing function that returns a region of an input image as its output.}

% 7. Using reinforcement learning for hard attention
The use of reinforcement learning to create hard attention has the following main benefits: (i) Removing the need for localization annotations in training data and using the currently available datasets. (ii) Removing the need for a separate attention network through the design of a multi-task network based on reinforcement learning. Annotations regarding action localization are not standard in action recognition and violence detection datasets. Through the conversion of video violence detection tasks to reinforcement learning tasks, the network is able to use a partial signal (classification annotation) within a reward system to simultaneously learn recognition and localization.

% 9. Human-machine interaction using gaze
% \hl{Furthermore, as a by-product of the hard attention implementation, the model is capable of rough violence localization. The attention window selected in the region selection phase approximately acts as a bounding box for the violence that is taking place in the video. The initial prior boxes chosen for the model are what determine the accuracy of localization. The model's approximate localization capability has a more significant meaning. The attention mechanism increases the model's action interpretability while increasing its efficiency. Visualizing the model's gaze in a real-time application can communicate what has captured the model's attention. This communication and clearness are integral for a more engaging human-AI interaction.}

\subsection{Design}
\label{3:2}

\begin{figure*}[!htbp]
    \centering
    \includegraphics[width=.9\textwidth]{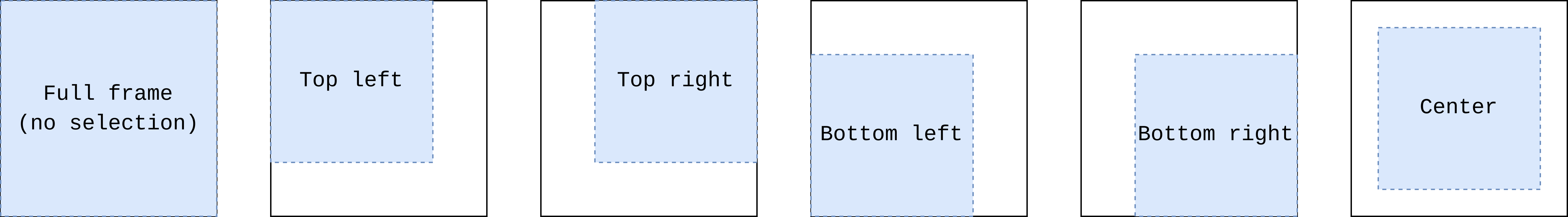}
    \caption{Prior boxes defined on the input frame.}
    \label{fig:priorbox}
\end{figure*}

\begin{figure*}[!htbp]
    \centering
    \includegraphics[width=.7\textwidth]{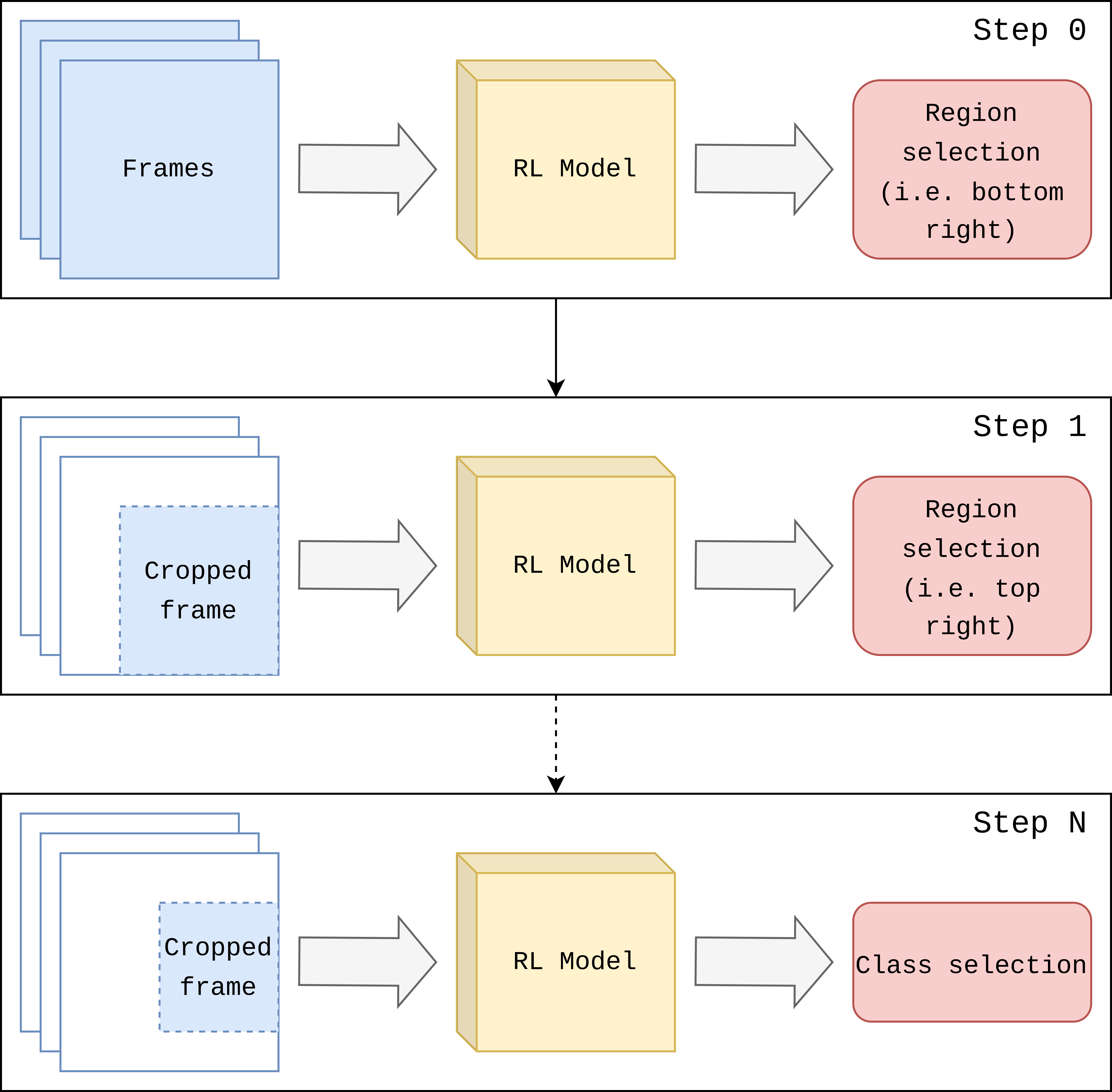}
    \caption{Model interaction with an input video.}
    \label{fig:interaction}
\end{figure*}

% 10. Multi-stage hard attention
%\hl
{The hard attention is implemented as a multi-stage process. The reinforcement learning environment utilizes predefined regions called prior boxes, as shown in} \autoref{fig:priorbox}%\hl
{. Through the use of prior boxes, the model can focus on different regions of a video within one or more region selection stages. In each stage, the chosen area replaces the current frame; as a result, the rest of the inference is carried out using the chosen region. In this manner, the model may continue to tighten the attention region by selecting the most appropriate area at each subsequent iteration. The purpose of the training is to teach the model how to select regions that contain individuals who exhibit violent behavior. Choosing a class (violent or non-violent) for the video concludes the violence recognition task. This process is depicted in} \autoref{fig:interaction}.

% \hl{The region selection and class selection tasks are done using a multi-task deep reinforcement learning network to evade the need for a separate attention and classification network. Consequently, the attention level (number of crops or zooms) is learned within the training process. The network learns to focus more (zoom more) when the region of interest is small and far and focus less when it is large and close.}

% 11. Single ROI and its extension
The implemented hard attention mechanism has only one region of interest at a given time. Although this may be a limitation, in the case of video violence detection, multiple regions of interest are not required. To correctly classify a video, it is sufficient to detect one of many acts of violence occurring within a single region. By transforming SSHA into a multi-agent reinforcement learning problem, SSHA can be extended to a larger number of attention regions. Attention agents work in collaboration to cover multiple areas of interest in order to maximize global precision.

% 12. Prior boxes vs free moving attention
Selection of the region is accomplished by choosing from a set of prior boxes. Using static regions rather than a free-moving attention bounding box in the training phase reduces the search space for the deep reinforcement learning model. As opposed to the limited number of predefined prior bounding boxes, a free-moving attention bounding box can possess all the possible bounding boxes on the input image. Due to the limited number of sample videos in the video violence recognition datasets, reducing the reinforcement learning search space enhances the convergence and generalization of the learned network weights.

% 13. Rewards (main and auxiliary)
%\hl
{A reward system complements the definition of a search space and an environment. The rewards are designed to encourage correct video classification and discourage incorrect classification. The rewards are defined to be +1 for a correct video class selection and -1 for an incorrect class selection. Furthermore, a diminishing +0.5 reward is associated with attention action in order to encourage the reinforcement learning model to experiment with region selection. As rewards below 0 act as punishments, the auxiliary reward should be greater than 0 to encourage the model. Additionally, since the reward for a correct class selection is +1, the auxiliary reward must not overwhelm the primary reward by being more than +1. According to empirical analysis, values such as 0.3, 0.4, 0.5, and 0.7 result in reasonably similar results, with 0.5 being the best.}

% 15. Inputs of the model
RGB and optical flow frames are used as raw inputs to the SSHA model. RGB frames are 224*224*3, and 79 RGB frames are sampled and fed to the model at each step. Optical flow input has similar dimensions, except the frames represent 2D motion vectors using two channels (instead of three RGB channels). The optical flow frames are calculated using the TV-L1 algorithm \citep{carreira2017quo}. TV-L1 is the algorithm of choice for the I3D backbone \citep{carreira2017quo} for its highly accurate motion vectors. \autoref{eq:tvl1} displays TV-L1 visual movement calculations.

\begin{equation}
\label{eq:tvl1}
\vec{v} = \displaystyle \min_{ \vec{u} } \left\{ \underbrace{\int_{\Omega}^{} \left| \nabla x \right| + \left| \nabla y \right|dt}_{\text{regularization term}} + \lambda \underbrace{\int_{\Omega}^{} \left|  \rho \left( x , y \right) \right|dt}_{\text{optical flow contraint}}  \right\}
\end{equation}

% 15.1 TVL1 equation description
%\hl
{By finding the smallest displacement vector ($\vec{u}$) in each region of an image, the motion vector ($\vec{v}$) can be calculated. In the unconstrained form, motion vectors are calculated as the displacement of a pixel within a short period of time ($\int_{/Omega}*{} \left| \nabla x \right| + \left| \nabla y \right| dt$). Where $\nabla x$ and $\nabla y$ are the displacement amounts of the pixel. The TVL1 optical-flow equation is constrained as it considers the weighted ($\lambda$) value difference of the tracked pixel. This difference is calculated as the derivative of the pixel value over a given period of time ($\left|  \rho \left( x , y \right) \right|dt$).}

% 14. Why use pre-trained backbone
The use of a pretrained model in this research improves the final accuracy, stabilizes deep reinforcement learning training, and accelerates model convergence. I3D model \citep{carreira2017quo} is chosen as the backbone network. Aside from being one of the best action recognition models on the leaderboards, this model also has excellent source code and pre-trained weights that make it suitable for the application. Kinetics dataset \citep{smaira2020short} is used to train the model's pre-trained weights, including RGB and Optical-flow streams.

\begin{figure*}[!htbp]
    \centering
    \includegraphics[width=0.9\textwidth]{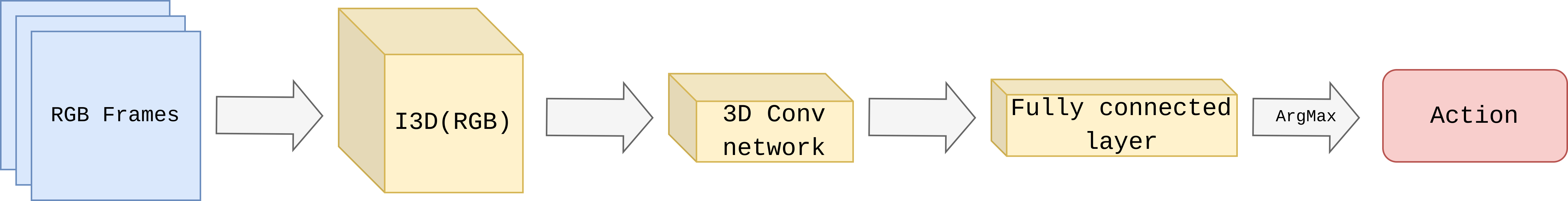}
    \caption{SSHA model architecture (RBG only).}
    \label{fig:architecture}
\end{figure*}

\begin{figure*}[!htbp]
    \centering
    \includegraphics[width=0.9\textwidth]{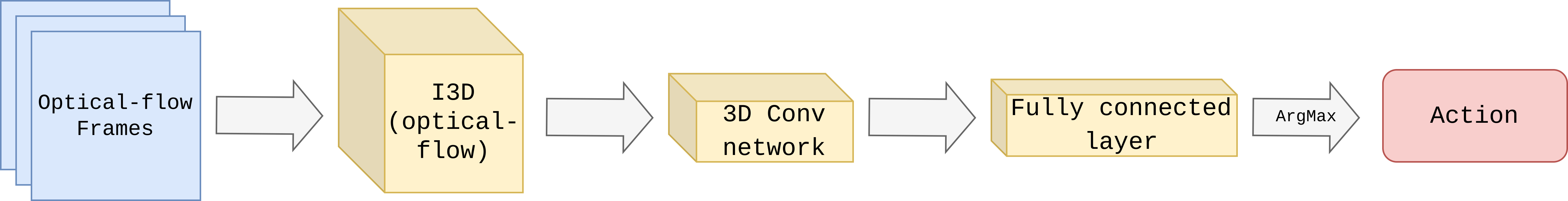}
    \caption{SSHA model architecture (Optical-flow only).}
    \label{fig:architecture_opt}
\end{figure*}

\begin{figure*}[!htbp]
    \centering
    \includegraphics[width=0.9\textwidth]{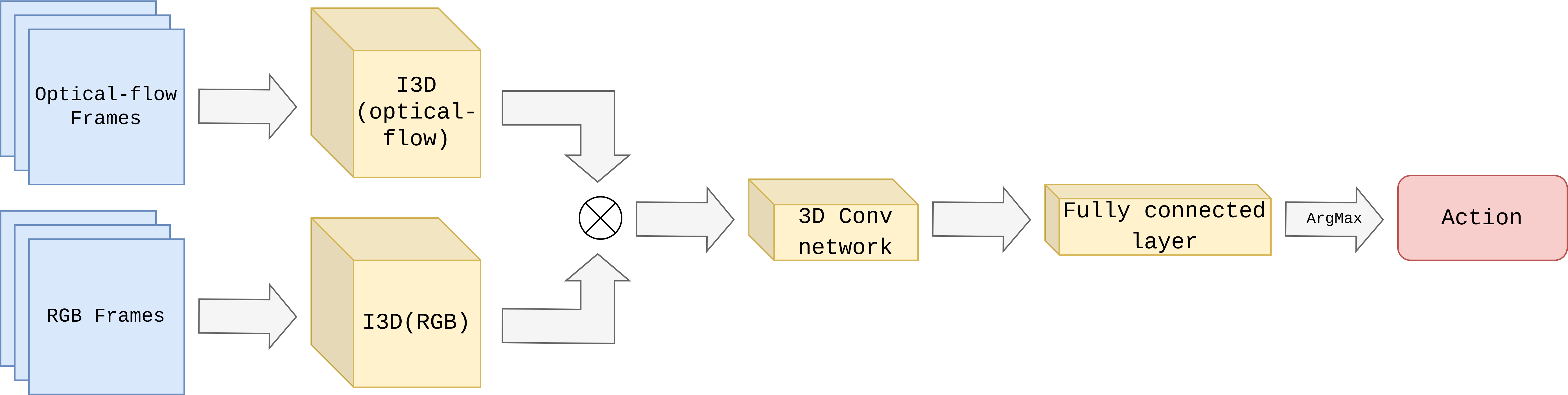}
    \caption{SSHA model architecture (Two-stream fusion).}
    \label{fig:architecture_twostream}
\end{figure*}

% 16. Multi-stream architecture + classification layers
To simultaneously use RGB and optical flow features, a two-stream architecture \citep{feichtenhofer2016convolutional} with a multiplication fusion layer is implemented. As shown in \autoref{fig:architecture_twostream}, the two-stream fusion network uses RGB and Optical-flow I3D backbones. The rest of the network has the same architecture as the single-stream configuration. The 3D feature map generated by the backbones is used to learn Q values. Feature maps are reduced in size using 3D convolution layers, which reduce the number of parameters and prevent over-fitting. The reduced feature is fed to the fully-connected layer with linear activations. Figures \ref{fig:architecture} to \ref{fig:architecture_twostream} demonstrate an overall view of various SSHA model architectures.

\subsection{Training}

% 17. Why Q learning
The SSHA model is trained using Q learning. A value iteration method (such as Q learning) uncovers the underlying value function of an environment. Comparatively, policy iteration approaches immediately identify the most effective actions for each state. The advantage of knowing the hidden value structure of the environment is the higher data efficiency {\citep{hamadouche2021comparison}}. In this study, value iteration methods are preferred because the dataset size is limited (e.g., 1600 training videos in the RWF dataset). The Q-learning equation is presented in {\autoref{eq:qlearning}}.

\begin{equation}
\label{eq:qlearning}
\underbrace{Q(s,a)}_{\scriptstyle\text{New Q value}}=\underbrace{R(s,a)}_{\scriptstyle\text{Reward}}+\mkern-30mu\underset{\text{Discount factor}}{\underset{\Biggl|}{\gamma}}\mkern-85mu\overbrace{\underset{a'}\max \, Q'(s',a')}^{\scriptstyle\substack{\text{Max expected Q value} \\ \text{obtained using the target network}}}
\end{equation}

% 18. Q learning equation description
%\hl
{Q value update given the current state and action ($Q(s, a)$) is calculated using} \autoref{eq:qlearning}%\hl
{. The new Q and observed reward ($R(s, a)$) are functions of current state ($s$) and the applied action ($a$). While the expected Q value ($Q'$) is the function of expected state ($s'$) and future action ($a'$).
}

% 19. Actions + finish criteria
The network outputs include two distinct groups of nodes that select attention regions or video classes. The node with the highest Q-value indicates the network choice at each stage. When a region selection action is selected, the intended region is cropped and used as input for the next inference step. In contrast to region selection actions, class selection actions indicate the classification decision and complete the violence detection process.

% 20. Step feature + diminishing reward for stepping
SSHA does not use recurrent or memory layers, so it cannot remember how many steps it took. In order to compensate for the lack of memory, the number of steps is fed into the fully connected layer. A one-hot vector encodes the number of steps. Making the network aware of the number of actions taken previously prevents the infinite loop scenario of selecting regions indefinitely.

% 21. Dueling network, e-greedy, adaptive sampling
In general, vanilla deep reinforcement learning models are unstable during training. To stabilize deep models, dueling training {\citep{wang2016dueling}} is used along with regularization, small learning rates, and batch-size tuning. Moreover, an adaptive sampling method maintains the reward sparsity in the $/epsilon - greedy$ exploration algorithm \citep{sutton2018reinforcement}. The reward sparsity is defined as the low probability of receiving positive rewards, while the model explores the environment in the training phase. The adaptive sampling method keeps the reward sparsity constant throughout the training phase by calculating the current reward sparsity and tuning the sampling probability from positive and negative rewards accordingly. Adaptive sampling is especially beneficial in the early stages of training as the model is entirely random, and environment complexity causes the samples to be primarily negative.

% 22. Reward injection
% 23. Reward clipping
The training is further stabilized and accelerated using a reward injection technique. Based on the known classification label for each video, we can determine the expected reward for correct and incorrect classifications. Since relevant information is provided without additional observations, the known reward accelerates the training. Since the output activation function is linear (pass-through) and the randomly initialized model outputs can have a large range. This strategy stabilizes the training phase by reducing the search space size at the start of the training procedure. Large Q values disrupt the main network's training stability because of the large learning gradients. The reward trimming method is also used to maintain training stability. Using this method, Q values that exceed or fall below a predefined range would be clipped to the maximum or minimum possible value. Applications and architectures define the predefined range. Considering the region selection reward, the expected Q value is between $-1$ and $1 + (N - 1) * .5$, where $N$ is the maximum number of steps the environment allows. In this study, N is equal to 5 as with more zooming, the quality of the input video falls below the SSHA model's input size. SSHA’s training procedure is thoroughly demonstrated in Algorithm \ref{alg:general} to \ref{alg:update}.

\begin{algorithm}
  \caption{SSHA training: train loop.}
  \label{alg:general}
  \begin{algorithmic}[1]
    \FOR{$num\_episodes$}
        \STATE $exploration()$
        \STATE $network\_update()$
    \ENDFOR
  \end{algorithmic}
\end{algorithm}

\begin{algorithm}
  \caption{SSHA training: $\epsilon -greedy$ exploration.}
  \label{alg:action}
  \begin{algorithmic}[1]
    \STATE $state_{i-1} = state_{i}$
    \STATE $action = -1$
    \WHILE{$action == -1$}
        \IF{$random\_number() < .5$}
            \IF{$random\_number() < \epsilon$}
                \STATE $selected\_action = select\_random\_action()$
            \ELSE
                \STATE $selected\_action = argmax(main\_network(state_{i}))$
            \ENDIF
            
            \STATE $state_{i}, reward, done = environment.act(action)$
        \ELSE
            \STATE $state_{i-1}, action, state_{i}, reward, done =$ \\ \hskip\algorithmicindent $replay\_buffer.random\_sample(action)$
        \ENDIF
        
        \STATE $prob = random\_number()$
        \IF{$(reward > 0$ \\ \hskip\algorithmicindent \AND $prob < positive\_reward\_selection\_prob)$ \\ \hskip\algorithmicindent \OR $(reward < 0 \ \AND \ prob >$ \\ \hskip\algorithmicindent $positive\_reward\_selection\_prob)$}
            \STATE \hskip\algorithmicindent $action = selected\_action$
            \STATE \hskip\algorithmicindent $reward\_history.append(reward)$
            \STATE \hskip\algorithmicindent {$replay\_buffer.append($ \\
            \hskip\algorithmicindent \hskip\algorithmicindent $state_{i-1}, action, state_{i}, reward, done)$}
        \ENDIF
    \ENDWHILE
  
  \end{algorithmic}
\end{algorithm}

\begin{algorithm}
  \caption{SSHA training: network update.}
  \label{alg:update}
  \begin{algorithmic}[1]
    \STATE $positive\_reward\_prob =$ \\ \hskip\algorithmicindent $calculate\_positive\_reward\_prob(reward\_history)$
    \IF{$positive\_reward\_prob > target\_positive\_reward\_prob$}
        \STATE $decrease\_positive\_reward\_selection\_prob()$
    \ELSE
        \STATE $increase\_positive\_reward\_selection\_prob()$
    \ENDIF
    \STATE
    \IF{$done$}
        \STATE $state_{i} = environment.reset()$
    \ENDIF
    \STATE
    \STATE {$Q_{target} = Q\_learning\_equation($ \\ \hskip\algorithmicindent $state_{i},\ reward,\ target\_network)$}
    \STATE $Q_{target} = inject\_known\_Q\_values(Q_{target})$
    \STATE $optimize\_main\_network(Q_{target})$
    \STATE $\epsilon = \epsilon - \frac{1}{num\_episodes}$
  \end{algorithmic}
\end{algorithm}

\section{Experiments}
\label{Experiments}
% \linenumbers

% 1. Experiments summary
Using classification metrics reported in the respective studies, the SSHA model is compared to the previous state-of-the-art models for video violence detection. Further, information regarding the class-level performance and model actions is provided for a more detailed assessment of the SSHA model. Based on the evaluation results, the advantages and limitations of the SSHA model are discussed in \autoref{Discussion}.

\subsection{Experiments setup}

% 2. Introducing the RWF dataset
The primary dataset used in this study is the RWF dataset \citep{cheng2021rwf}. The RWF dataset is one of the most comprehensive datasets available for the detection of video violence. A total of 2000 videos were included in the RWF dataset, divided into violent and non-violent categories of equal size. Video content in RWF is more abundant in quantity and diversity than in previously published datasets. Because this dataset is scraped from the YouTube \footnote{https://youtube.com} video streaming service, it contains videos with varying resolutions. The videos are divided into five-second clips, each with a frame rate of 30 frames per second. It is easier to compare competing models with the help of a predefined list of train and test videos. Classification accuracy is the reported metric in the baseline RWF paper is \citep{cheng2021rwf}.

% 3. limitations of RWF dataset
% While the RWF dataset is an excellent starting point for violence detection research, there is still much room for improvement. The suggested improvements are related to the quality and quantity of the dataset. Quality-wise, a portion of this dataset ($\sim$6\%) is noisy and of low quality. By feeding noisy information to a neural network, the accuracy of the model is reduced. Furthermore, a small percentage of videos ($/sim$1/\%) do not meet the minimum frame rate requirement of 30 frames per second. The different timing of frames in these videos compared to the rest of the dataset misleads the optimization algorithm into choosing non-optimum 3D filters. Lastly, this dataset lacks a complete representation of various types of violence, such as stabbings, shootings, and chases.

% 4. Introducing moves and hockey datasets
Hockey and movie fight datasets are used in addition to the RWF for SSHA model evaluation. Despite containing 1000 and 200 videos, they do not meet the criteria for a practical dataset for building a video violence detection model in the real world. In these datasets, violence is only depicted in hockey games and Hollywood films. A generalizable violence detection model cannot be trained with the repeating situation of hockey players fighting on an icy hockey field and the cinematic quality and perspective of the cinematic film. Therefore, despite the high accuracy of models on the mentioned datasets, they are not useful for real-world and general applications. Dataset characteristics are presented in \autoref{table:datasets}.

\begin{table}[!htbp]
\caption{Video violence datasets characteristics. }
\label{table:datasets}
\resizebox{\columnwidth}{!}{
\begin{tabular}{lccc}

\textbf{Dataset} & \textbf{\# videos} & \textbf{Video length (seconds)} & \textbf{Video size (pixels)} \\

RWF \citep{cheng2021rwf} & 2000 & 5 & varied \\
Hockey fights \citep{nievas2011violence} & 1000 & 2 & 360 x 288 \\
Movie fights \citep{gong2008detecting} & 200 & 2 & 720 x 480 \\

\end{tabular}
}
\end{table}

\subsection{Results}

% 5. Train/test split
The SSHA model is trained and tested using the predefined training and testing videos included in the RWF dataset. Furthermore, the hockey and movie datasets are divided 80/20 between training (80\%) and testing (20\%). SSHA's accuracy and class-level performance are presented in \autoref{table:models} and \ref{table:classlevel}, respectively. The class-level results provide insight into the SSHA model's performance in detecting violent and non-violent videos in isolation. In addition, \autoref{table:statistics} provides some characteristics of the SSHA model, such as the model size and the average number of actions per video.

\begin{table}[!htbp]
\caption{Models accuracy banchmark. }
\label{table:models}
\resizebox{\columnwidth}{!}{
\begin{tabular}{lccc}

\textbf{Model/Dataset)}   & \textbf{RWF \citep{cheng2021rwf}} & \textbf{Hockey fights \citep{nievas2011violence}} & \textbf{Movie fights \citep{gong2008detecting}} \\

ConvLSTM \citep{sudhakaran2017learning} & 77.0 \% & 97.1 \% & \textbf{100.0} \% \\
I3D (RGB only) \citep{carreira2017quo} & 85.7 \% & 98.5 \% & \textbf{100.0} \% \\
I3D (Optical-flow only) \citep{carreira2017quo} & 75.5 \% & 84.0 \% & \textbf{100.0} \% \\
I3D (Two-stream) \citep{carreira2017quo} & 81.5 \% & 97.5 \% & \textbf{100.0} \% \\
Cheng et al. (RBG only) \citep{cheng2021rwf} & 84.5 \% & - & - \\
Cheng et al. (Optical-flow only) \citep{cheng2021rwf} & 75.5 \% & - & - \\
Cheng et al. (C3D) \citep{cheng2021rwf} & 85.7 \% & - & - \\
Cheng et al. (P3D) \citep{cheng2021rwf} & 87.2 \% & 98.0 \% & \textbf{100.0} \% \\
SSHA model (RBG only no localization) & 85.3 \% & 98.0 \% & 99.0 \% \\
SSHA model (RGB only) & \textbf{90.4} \% & \textbf{98.7} \% & 99.0 \% \\
SSHA model (Optical-flow only) & 76.0 \% & 86.2 \% & 98.5 \% \\
SSHA model (Two-stream) & 86.4 \% & 97.0 \% & 99.0 \% \\

\end{tabular}
}
\end{table}

\begin{table}[!htbp]
\centering
\caption{SSHA Class-level evluation results on RWF dataset. }
\label{table:classlevel}
\resizebox{.6\columnwidth}{!}{
\begin{tabular}{lccc}

\textbf{Class}   & \textbf{Precision} & \textbf{Recall} & \textbf{F1-score} \\

Violent & 0.88 \% & 0.92 \% & 0.9 \% \\
Non violent & 0.92 \% & 0.9 \% & 0.91 \% \\
Total & 0.9 \% & 0.91 \% & 0.9 \% \\

\end{tabular}
}
\end{table}

\begin{table}[!htbp]
\centering
\caption{SSHA Class-level evluation results. }
\label{table:statistics}
\resizebox{.6\columnwidth}{!}{
\begin{tabular}{lc}

\textbf{Property}   & \textbf{Value} \\

\# parameters (total) & 13.2 million \\
\# parameters (trained) & .9 million \\
\# input frames & 79 \\
Input size & 224 x 224 pixels \\
Avg. \# actions per video (RWF) & 1.8 \\

\end{tabular}
}
\end{table}

\section{Discussion}
\label{Discussion}
% \linenumbers

% 0. Might need a starting paragraph
As shown in \autoref{table:models} the SSHA model has achieved state-of-the-art accuracy on RWF and Hockey fights datasets, and fair accuracy on the Movie fights dataset using the RBG-only architecture. The accuracy of models has reached saturation for the Hockey and Movie fights datasets, but state-of-the-art models continue to produce meaningful results on RWF. As a result, the SSHA model's advantages are more evident in the RWF dataset. Furthermore, this study does not suffer from imbalance learning because the evaluation datasets are perfectly balanced. As shown in \autoref{table:classlevel}, the class-level results of the SSHA model on the RWF dataset present a proper balance between the violence and non-violence classes.

% 2. Localization-less evaluation
To assess the effects of hard attention on the accuracy of the SSHA model, the SSHA model is trained and evaluated with and without hard attention capability (RGB only and RGB only no localization in \autoref{table:models}). The attention mechanism is detached from the SSHA model by removing the region selection actions from the model and converting it to a single-stage video classification model. Nonetheless, this model is still trained using reinforcement learning loss. The superior accuracy of the SSHA model with region selection capabilities demonstrates the effectiveness of the hard attention method.

% 4. Action characteristics
Learned from the training phase, 1.8 is the optimized average number of actions per video learned by the SSHA model (including the classification action). When only one step is taken to classify a video, the model has classified the video without applying region selection. This scenario is reasonable when the region of interest is large; thus, no further attention is required for accurate classification. Many region selection actions result in a narrow viewpoint on the input video. A perspective with such a narrow field of view often lacks the necessary visual details. Consequently, the average number of actions per video indicates the model’s tendency toward one region selection action before classification.

% 1. Comparison of RGB only, Optical-flow only, and Two stream architectures
The notable conclusion of experimenting with a two-stream fusion neural architecture is the drawback of having a larger model trained on a limited dataset. Even though an I3D network with a two-stream fusion architecture achieves higher accuracy on the Kinetics dataset \citep{carreira2017quo}, results on the RWF dataset do not follow the same principle. According to \autoref{table:models}, the I3D two-stream fusion has inferior accuracy on the RWF dataset compared to the RGB-only type in previous and current research. The bad performance results from expanding the feature space and, subsequently, the neural network search space while having a fixed number of training data samples. However, the positive aspect of this outcome is the performance-wise preferability of an RGB-only architecture. The I3D backbone has 13 million parameters \citep{carreira2017quo}. A two-stream architecture utilizes two I3D backbones for feature extraction from RGB and Optical-flow frames, doubling the number of parameters relative to the single-stream architecture. Thus, the computational overhead of extracting Optical-flow frames from RGB frames and a double-sized neural network adversely affects the two-stream architecture’s performance.

\section{Conclusion}
\label{Conclusion and future work}
% \linenumbers

%\hl
{
% 1. Conclusions
This paper presents a semi-supervised hard attention mechanism (SSHA) based on reinforcement learning. SSHA achieves state-of-the-art accuracy in video violence detection by analyzing the most critical region of the video in greater depth. It utilizes video violence datasets that are readily available and eliminates the need for specialized datasets or annotations. The multi-stage implementation of SSHA enables the proposed model to utilize high-definition surveillance footage by selecting attention regions according to the user's preferences. The RGB-only version of the SSHA model achieved state-of-the-art 90.4 percent accuracy on the RWF dataset and 98.5 and 99.5 percent accuracy on the Hockey and Movies datasets, respectively.

% 2. Future work
Even though the proposed SSHA model significantly improved the accuracy of the existing state-of-the-art models, future research could apply the hard attention mechanism to action recognition to improve the accuracy of SSHA methods. Additionally, applying the proposed hard attention mechanism to multi-attention scenarios using collaborative agents would be an interesting future direction for this research. Contributing to the RWF dataset regarding the number and quality of available videos and annotations will provide the greatest benefit to state-of-the-art automated video violence detection in the short term.
}

\bibliography{refs}

\end{document}